\pdfoutput=1

\documentclass[11pt]{article}

\usepackage[]{ACL2023}

\usepackage{times}
\usepackage{latexsym}

\usepackage[T1]{fontenc}

\usepackage[utf8]{inputenc}

\usepackage{microtype}

\usepackage{inconsolata}

\usepackage{graphicx}
\usepackage{amsfonts,amssymb}
\usepackage{amsmath}
\usepackage{multirow}
\usepackage{arydshln}

\title{RC$^{3}$: Regularized Contrastive Cross-lingual Cross-modal Pre-training}

\author{
  Chulun Zhou\textsuperscript{1}\footnotemark[1]  ,
  Yunlong Liang\textsuperscript{2}\footnotemark[1]  ,
  Fandong Meng\textsuperscript{1}\footnotemark[2], 
  \textbf{Jinan Xu}\textsuperscript{2}, 
  \textbf{Jinsong Su}\textsuperscript{3}
   and \textbf{Jie Zhou}\textsuperscript{1}\\
  \textsuperscript{1}Pattern Recognition Center, WeChat AI, Tencent Inc, China \\
  \textsuperscript{2}Beijing Key Lab of Traffic Data Analysis and Mining, \\Beijing Jiaotong University, Beijing, China \\
  \textsuperscript{3}School of Informatics, Xiamen University, Xiamen, China \\
  \texttt{\{chulunzhou,fandongmeng,withtomzhou\}@tencent.com} \\
  \texttt{\{yunlongliang,jaxu\}@bjtu.edu.cn} \\
  \texttt{jssu@xmu.edu.cn} \\
}

\begin{document}
\maketitle
\renewcommand{\thefootnote}{\fnsymbol{footnote}}
\footnotetext[1]{Equal contribution.}
\footnotetext[2]{Fandong Meng is the corresponding author.}
\renewcommand{\thefootnote}{\arabic{footnote}}
\begin{abstract}
Multilingual vision-language (V\&L) pre-training has achieved remarkable progress in learning universal representations across different modalities and languages. In spite of recent success, there still remain challenges limiting further improvements of V\&L pre-trained models in multilingual settings. Particularly, current V\&L pre-training methods rely heavily on strictly-aligned multilingual image-text pairs generated from English-centric datasets through machine translation. However, the cost of collecting and translating such strictly-aligned datasets is usually unbearable. In this paper, we propose \textbf{R}egularized \textbf{C}ontrastive \textbf{C}ross-lingual \textbf{C}ross-modal (RC$^3$) pre-training, which further exploits more abundant weakly-aligned multilingual image-text pairs. Specifically, we design a regularized cross-lingual visio-textual contrastive learning objective that constrains the representation proximity of weakly-aligned visio-textual inputs according to textual relevance. Besides, existing V\&L pre-training approaches mainly deal with visual inputs by either region-of-interest (ROI) features or patch embeddings. We flexibly integrate the two forms of visual features into our model for pre-training and downstream multi-modal tasks. Extensive experiments on 5 downstream multi-modal tasks across 6 languages demonstrate the effectiveness of our proposed method over competitive contrast models with stronger zero-shot capability.
\end{abstract}

\begin{figure}[t]
    \centering
    \includegraphics[width=0.45\textwidth,height=0.16\textheight]{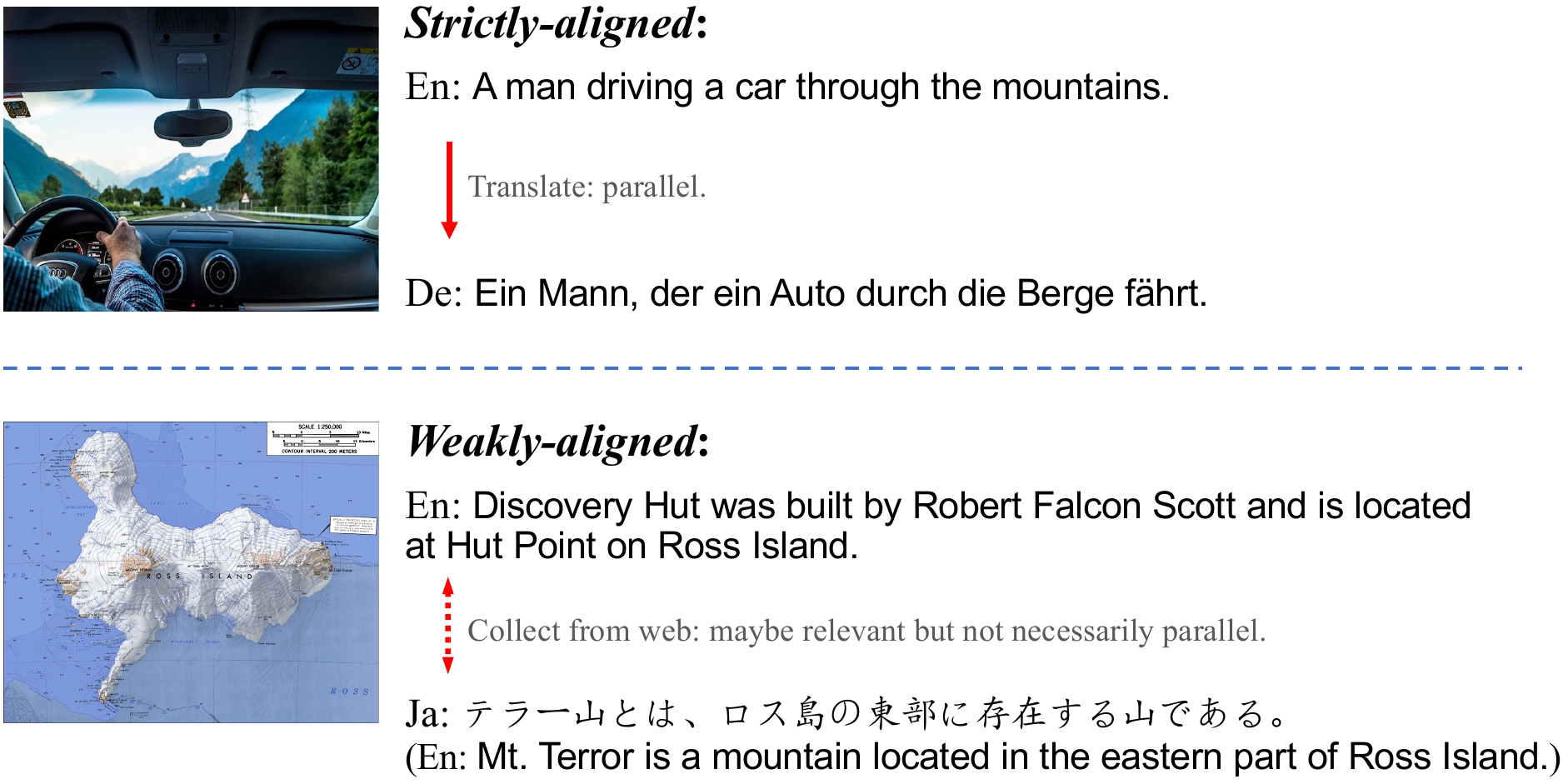}
    \caption{Comparison between ``strictly-aligned'' and ``weakly-aligned'' image-text pairs in different languages.}
    \label{fig:alignment}
    \vspace{-10pt}
\end{figure}
\section{Introduction}
Vision-language (V\&L) pre-training aims to learn universal representations that can express visual and textual semantics informatively. It exploits a large amount of multi-modal data (\textit{e.g.} image-text pairs) to make the model capable of handling cross-modal data. Till now, the advents of various V\&L pre-trained models have achieved remarkable results on many downstream multi-modal tasks. 

Recently, V\&L pre-trained models have developed from focusing on English-dominant tasks \cite{DBLP:conf/iclr/SuZCLLWD20,DBLP:conf/eccv/ChenLYK0G0020,DBLP:conf/icml/ChoLTB21} into multilingual scenarios \cite{DBLP:conf/cvpr/NiHSCBW0D21,DBLP:conf/emnlp/0001BPRCE21,DBLP:conf/cvpr/ZhouZW0LYL21}. To this end, researchers construct multi-modal data in multiple languages and design various cross-lingual pre-training objectives. Such advances enable multi-modal modelling to leverage more diverse language resources. Meanwhile, these multilingual V\&L pre-trained models also show their advantages over previous English-centric models in terms of generalization abilities across languages, especially in zero-shot settings.

Despite the promising performances of current multilingual V\&L models, one of the major challenges is that they usually require massive strictly-aligned multilingual image-text pairs. The prevalent practice is to translate English-only multi-modal datasets into pseudo-parallel multilingual versions via machine translation (MT) \cite{DBLP:conf/cvpr/NiHSCBW0D21,DBLP:conf/cvpr/ZhouZW0LYL21}. However, the cost of collecting and translating such large-scale multi-modal datasets is often unbearable. To deal with this issue, we turn our eyes on those more easily available weakly-aligned multilingual multi-modal data, such as WIT \cite{DBLP:conf/sigir/Srinivasan0CBN21}. As shown in Figure~\ref{fig:alignment}, the so-called ``weakly-aligned'' means that the multilingual textual data of the same image are not strictly parallel.

In this paper, we propose a \textbf{R}egularized \textbf{C}ontrastive \textbf{C}ross-lingual \textbf{C}ross-modal (RC$^{3}$) pre-training framework, which can make better use of relatively abundant weakly-aligned multilingual image-text pairs. Specifically, we adopt an encoder-decoder architecture so that our model can be more adaptive to both discriminative and generative downstream tasks. Besides the widely used image-text matching (ITM) task, we further introduce masked conditional language modelling (MCLM) and cross-lingual textual contrastive learning (XTCL) along with our proposed regularized cross-lingual visio-textual contrastive learning (R-XVtCL) during pre-training. Particularly, while R-XVtCL encourages the visio-textual representations of two weakly-aligned image-text pairs to be close, a regularization term is designed to constrain such proximity according to the textual relevance of their respective texts.

Meanwhile, in current V\&L models, there are mainly two ways of processing visual inputs:(1) Region-of-interest based (ROI-based). It uses external object detectors (\textit{e.g.} Faster-RCNN~\cite{DBLP:conf/nips/RenHGS15}) to extract ROI features from images and feed them with paired text into V\&L models \cite{DBLP:conf/iclr/SuZCLLWD20,DBLP:conf/eccv/ChenLYK0G0020,DBLP:conf/icml/ChoLTB21,DBLP:conf/cvpr/NiHSCBW0D21,DBLP:conf/emnlp/0001BPRCE21,DBLP:conf/cvpr/ZhouZW0LYL21}. This method exerts the informativeness of ROI features, but such cumbersome protocol hinders the usage of massive online image-text pairs and requires additional procedures for various downstream tasks. (2) Patch-based. It directly transforms the original image pixels into patch embeddings and take them as inputs with textual data \cite{DBLP:conf/icml/JiaYXCPPLSLD21,DBLP:conf/coling/LeeLBRK22,DBLP:conf/iclr/WangYYDT022}. This significantly simplifies pre-training protocols but cannot leverage informative ROI features. To improve the informativeness of visual features without complicating the whole training protocol, we flexibly integrate the above two forms of visual features into the model for pre-training and downstream tasks.

Our contributions can be summarized as follows: (1) We propose a cross-lingual cross-modal pre-training framework that can better exploit more abundant weakly-aligned multilingual image-text pairs; (2) We integrate ROI-based and patch-based visual features to enhance our V\&L model for pre-training and downstream multi-modal tasks; (3) Extensive experiments on 5 downstream tasks across 6 languages show that our V\&L model achieves higher or comparable performances over recent competitive contrast models with strong zero-shot capability.
\section{Our Approach}
In this section, we first briefly introduce the three types of datasets used for pre-training and more details are given in Appendix~\ref{sec:pretrain_datasets}. Then, we describe the model architecture and pre-training objectives.
\subsection{Pre-training Data}
\label{subsection:pretrain_data}
\paragraph{\textbf{Strictly-aligned Multilingual Image-caption Dataset $D_{s}$.}}
We use the machine translation augmented image-caption paired data released in \cite{DBLP:conf/cvpr/ZhouZW0LYL21}. The English captions from Conceptual Captions dataset \cite{DBLP:conf/acl/SoricutDSG18} are translated into five different languages (Czech, German, French, Japanese and Chinese). This gives rise to a final strictly-aligned multilingual visio-linguistic dataset $D_{s}$, each image of which is paired with semantically-equivalent captions of 6 languages. 
\paragraph{\textbf{Weakly-aligned Multilingual Image-text Dataset $D_{w}$.}}
We build a weakly-aligned visio-linguistic dataset $D_{w}$ by extracting a fraction of multilingual image-caption pairs of 6 languages (German, English, French, Indonesian, Japanese and Chinese) from WIT dataset \cite{DBLP:conf/sigir/Srinivasan0CBN21}. Note that the attached multilingual texts of the same image in $D_{w}$ are not strictly parallel. 
\paragraph{\textbf{Multilingual Parallel Text Dataset $D_{t}$.}}
We also use a combination of different textual data to form a multilingual parallel text dataset $D_{t}$. It is comprised of the parallel text corpus collected by \cite{DBLP:journals/corr/abs-2206-00621} from a subset of WikiMatrix  \cite{DBLP:conf/eacl/SchwenkCSGG21} and the parallel captions from $D_{s}$, which includes all 7 languages involved in $D_{s}$ and $D_{w}$ (\textit{i.e.} English, Czech, German, French, Indonesian, Japanese and Chinese). 

\subsection{Model Architecture}
\label{subsec:model_arch}
\begin{figure}[t]
    \centering
    \includegraphics[width=0.45\textwidth]{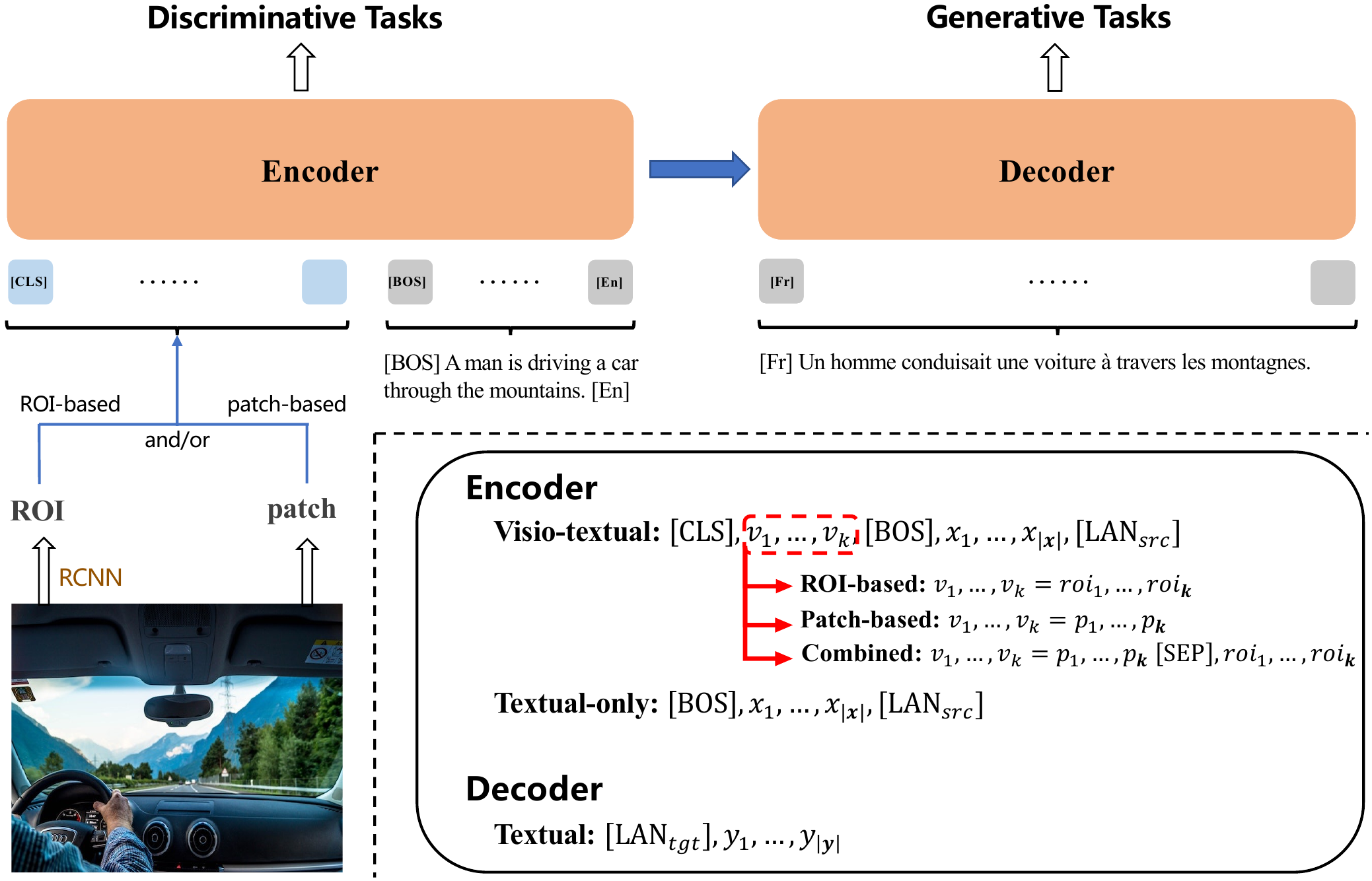}
    \caption{The architecture of our model and the sequence formats for visio-texual/textual-only inputs.}
    \label{fig:model_arch}
    \vspace{-10pt}
\end{figure}

We extend the encoder-decoder structure to make our model adaptive to both discriminative and generative multi-modal tasks. Figure~\ref{fig:model_arch} depicts the model architecture and the sequence formats for visio-textual/textual-only inputs.

\paragraph{Cross-lingual Cross-modal Encoder.}
As shown in Figure~\ref{fig:model_arch}, given a visio-textual input composed of an image and texts, the visual features are concatenated with text embeddings, which are then fed to the multi-layer encoder. Specifically, the visual features can be presented in the following three forms: (1) \textit{\textbf{ROI-based}}. The ROI features $\boldsymbol{roi}$ = $\{roi_1, roi_2, ..., roi_k\}$ generated from an external object detector are projected by a fully-connected (FC) layer to have the same dimension as text embeddings; (2) \textit{\textbf{Patch-based}}. Raw pixels are also mapped by another FC layer into a patch embedding sequence $\boldsymbol{p}$ = $\{p_1, p_2, ..., p_k\}$; (3) \textit{\textbf{Combined}}. To enhance the informativeness of visual features, ROI features and patch embeddings are combined and fed to the encoder together, between which a special token $\rm [SEP]$ is inserted. For texts, we add a special token $\rm [BOS]$ and a language tag. Finally, a special token $\rm [CLS]$ is prepended at the beginning of the concatenated sequence, the output hidden state of which serves as its visio-textual representation (VtR).

For a textual-only input, only text embeddings are fed to the encoder and the output hidden state corresponding to $\rm [BOS]$ is used as its textual representation (TR). 

\paragraph{Multilingual Decoder.}
In generative tasks that involve multiple languages, we also prepend a special language tag on the decoder side, indicating to which language the decoder is expected to generate texts.

\subsection{Pre-training Objectives}
\label{subsec:pretrain_objectives}
During training, we adopt four pre-training tasks: (1) Masked Conditional Language Modelling (MCLM); (2) Image Text Matching (ITM); (3) Cross-lingual Textual Contrastive Learning (XTCL); (4) Regularized Cross-lingual Visio-textual Contrastive Learning (R-XVtCL). These tasks train the model to capture cross-lingual cross-modal alignments among images and multilingual texts using different types of pre-training data described in Section~\ref{subsection:pretrain_data}.
\subsubsection{Masked Conditional Language Modelling (MCLM)}
Masked language modelling (MLM) has been widely used in previous encoder-only visio-linguistic models. Given an image $\boldsymbol{v}$ and its caption $\boldsymbol{x}^{l_i}$ in language $l_i$ from the strictly-aligned dataset $D_{s}$, a word in $\boldsymbol{x}^{l_i}$ has a probability of 15\% to be replaced with a special token [MASK]. The objective is to predict a set of masked words $\boldsymbol{x}^{l_i}_{m}$ based on other unmasked words $\boldsymbol{x}^{l_{i}}_{\backslash m}$ and the visual input:
\begin{equation}
    L_{MLM} = -\mathbb{E}_{(\boldsymbol{v}, \boldsymbol{x}^{l_i})\sim D_{s}} \log{P_{\theta_{e}}(\boldsymbol{x}^{l_i}_{m}|\boldsymbol{x}^{l_i}_{\backslash m}, \boldsymbol{v})},
\label{eq:mlm_objective}
\end{equation}
where $\theta_{e}$ is the trainable parameters of the encoder.

Moreover, with respect to $\boldsymbol{x}^{l_i}$, since $D_{s}$ also provides the parallel caption $\boldsymbol{x}^{l_j}$ in another language $l_j$, we simultaneously train the decoder to autoregressively predict the target text $\boldsymbol{x}^{l_j}$ based on the unmasked words $\boldsymbol{x}^{l_i}_{\backslash m}$ and $\boldsymbol{v}$. The MCLM objective can be formulated as follows:
\begin{align}
&L_{MCLM} = L_{MLM} \\
-&\mathbb{E}_{(\boldsymbol{v}, \boldsymbol{x}^{l_i}, \boldsymbol{x}^{l_j})\sim D_{s}} \sum_{t=1}^{|\boldsymbol{x}^{l_j}|}{\log{P_{\theta_{d}}(x^{l_j}_{t}|x^{l_j}_{<t},\boldsymbol{x}^{l_i}_{\backslash m}, \boldsymbol{v})}}, \notag
\label{eq:mclm_objective}
\end{align}
where $\theta_{d}$ is the trainable parameters of the decoder. In addition to MLM, the incorporation of the autoregressive term on the decoder can make the model better adapt to downstream generative tasks.

\subsubsection{Image Text Matching (ITM)}
ITM aims to discriminate whether an image and a piece of caption are matched, training the model to learn the alignment between visual and textual modalities. The representation of a visio-textual input $(\boldsymbol{v}, \boldsymbol{x}^{l})$ is fed to an FC layer and a sigmoid function to get a score $s_{\theta_e}(\boldsymbol{v}, \boldsymbol{x}^{l})$. The score ranges from 0 to 1, predicting to what extent $\boldsymbol{v}$ and $\boldsymbol{x}^{l}$ are matched.
We sample positive and negative visio-textual inputs from the strictly-aligned dataset $D_s$, where the negative one is constructed by randomly selecting another caption within the same batch to be paired with the original image. Thus, the training objective of ITM is written as
\begin{align}
L_{ITM}\,=\,-\mathbb{E}_{(\boldsymbol{v}, \boldsymbol{x}^{l})\sim D_{s}}& [y\log{s_{\theta_e}(\boldsymbol{v}, \boldsymbol{x}^{l})} \\
+(1-y)&\log{(1-s_{\theta_e}(\boldsymbol{v}, \boldsymbol{x}^{l}))}], \notag
\end{align}
where $y$ $\in$ $\{0,1\}$ indicates whether $(\boldsymbol{v}, \boldsymbol{x}^{l})$ is a negative or positive sample.

\begin{figure}[t]
    \centering
    \includegraphics[width=0.47\textwidth]{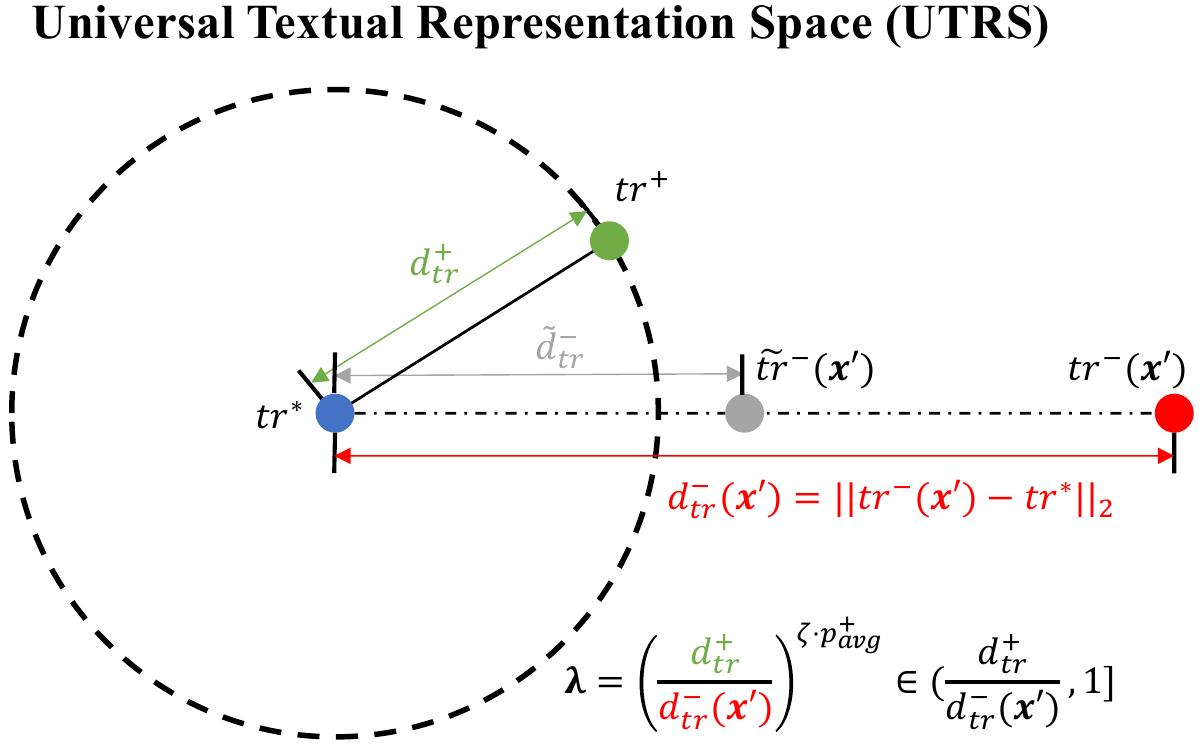}
    \caption{The construction of harder negative samples by smoothed linear interpolation. The blue and green points represent the anchor instance and the positive sample in the UTRS, respectively. The red point refers to a negative sample whose interpolated harder TR representation corresponds to the grey one.}
    \label{fig:linear_smooth_interpolation}
    \vspace{-10pt}
\end{figure}

\subsubsection{Cross-lingual Textual Contrastive Learning (XTCL)}
\label{subsec:XTCL}
XTCL is to learn semantically informative representations of multilingual texts in a universal textual representation space (UTRS), where the TR representation of semantically equivalent texts are expected to be close while those of irrelevant ones are far from each other. Therefore, we adopt the interpolation-based contrastive learning method introduced in \cite{DBLP:conf/iclr/WeiW0XYL21} to train the model, as depicted in Figure~\ref{fig:linear_smooth_interpolation}. 

Specifically, given a batch of parallel text pairs $B_t$ = $\{(\boldsymbol{x}^{l_{i}}_b, \boldsymbol{x}^{l_{j}}_b)\}_{b=1}^{|B_t|}$ ($l_{i}$$\neq$$l_{j}$) from the multilingual parallel dataset $D_t$, for a pair of parallel texts $(\boldsymbol{x}^{l_{i}}, \boldsymbol{x}^{l_{j}})$ $\in$ $B_t$, we treat $\boldsymbol{x}^{l_{i}}$ as the anchor textual instance, the representation of which serves as the anchor point $tr^{*}$ (the blue center) in the UTRS. Intuitively, the semantically equivalent $\boldsymbol{x}^{l_{j}}$ is naturally the positive sample and its representaion $tr^{+}$ (the green point on the circle) should be near to $tr^*$. On the contrary, each of the other texts $\boldsymbol{x}'$ within $B$ is used as the negative sample whose TR representation $tr^{-}(\boldsymbol{x}')$,  \textit{i.e.} the red point out of the circle, should be far from the anchor. The XTCL objective can be defined as
\vspace{-5pt}
\begin{equation}
\resizebox{0.98\hsize}{!}{$L_{xltcl}(\boldsymbol{x}^{l_i})=-\log{\frac{\exp{(-d^{+}_{tr})}}{\exp{(-d^{+}_{tr})}+\sum\limits_{\boldsymbol{x}' \in \mathcal{N}(\boldsymbol{x}^{l_{i}})}{\exp{(-d^{-}_{tr}(\boldsymbol{x}'))}}}}$},
\label{eq:loss_xltcl}
\end{equation}
where $\mathcal{N}(\boldsymbol{x}^{l_i})$ is the set of negative samples with respect to $\boldsymbol{x}^{l_i}$, $d^+_{tr}$ and $d^-_{tr}(\boldsymbol{x})$ denote the euclidean TR distances from $tr^{+}$ and each $tr(\boldsymbol{x}'^{-})$ to the anchor in the UTRS, \textit{i.e.} $d^+_{tr}$ = $||tr^{+}-tr^{*}||_2$ and $d^-_{tr}(\boldsymbol{x}')$ = $||tr^{-}(\boldsymbol{x}')-tr^{*}||_2$.

However, since the above negative samples are usually not informative, following \cite{DBLP:conf/iclr/WeiW0XYL21}, we generate harder negative samples by smoothed linear interpolation~\cite{DBLP:conf/conll/BowmanVVDJB16,DBLP:conf/cvpr/ZhengCL019}. For a negative sample $\boldsymbol{x}'$ from $N_{tr}$, a more difficult negative representation in the UTRS is constructed through the following interpolation:

\vspace{-10pt}
\begin{small}
\begin{equation}
\label{eq:tr_interpolation}
\widetilde{tr}^{-}(\boldsymbol{x}')=\left\{
\begin{aligned}
&tr^{*}+\boldsymbol{\lambda}(tr^{-}(\boldsymbol{x}')-tr^{*}), d^{-}_{tr}(\boldsymbol{x}')>d^{+}_{tr}; \\
&tr^{-}(\boldsymbol{x}'),\qquad\qquad\qquad d^{-}_{tr}(\boldsymbol{x}')\leq d^{+}_{tr};
\end{aligned}
\right.
\end{equation}
\end{small}
\vspace{-5pt}
\begin{equation}
\resizebox{0.40\hsize}{!}{$\boldsymbol{\lambda} = \left(\frac{d^{+}_{tr}}{d^{-}_{tr}(\boldsymbol{x}')}\right)^{\zeta \cdot p^{+}_{avg}},$}
\label{eq:lambda}
\end{equation}
where $p^{+}_{avg}$=$\frac{1}{100}\sum_{\tau\in[-100,-1]}{e^{-L_{xltcl}^{(\tau)}}}$ is the average log-probability over the previous 100 training steps in Equation~\ref{eq:loss_xltcl} and $\zeta$ is a slacking coefficient set to 0.9 in our experiment. By doing so, the difficulty of the interpolated representation $\widetilde{tr}^{-}(\boldsymbol{x}')$, \textit{i.e.} the grey point out of the circle in Figure~\ref{fig:linear_smooth_interpolation}, can be dynamically adjusted during training, which results in a lower $\boldsymbol{\lambda}$ (harder samples) when $p^+_{avg}$ increases and vice versa. 

Thus, Equation~\ref{eq:loss_xltcl}
is reformulated by replacing the original representation of each negative sample $\boldsymbol{x}'$ with the harder interpolated one $\widetilde{tr}^{-}(\boldsymbol{x}')$:
\begin{equation}
\resizebox{0.98\hsize}{!}{$\Tilde{L}_{xltcl}(\boldsymbol{x}^{l_i})=-\log{\frac{\exp{(-d^{+}_{tr})}}{\exp{(-d^{+}_{tr})}+\sum\limits_{\boldsymbol{x}' \in \mathcal{N}(\boldsymbol{x}^{l_i})}{\exp{(-\Tilde{d}^{-}_{tr}(\boldsymbol{x}'))}}}}$},
\label{eq:loss_xltcl_new}
\end{equation}
where $\Tilde{d}^{-}_{tr}(\boldsymbol{x}')$ is the euclidean distance between the anchor and $\widetilde{tr}^{-}(\boldsymbol{x}')$, \textit{i.e.} $\Tilde{d}^{-}_{tr}(\boldsymbol{x}')$=$||\widetilde{tr}^{-}(\boldsymbol{x}')-tr^{*}||_2$. Finally, the XTCL objective is:
\begin{equation}
\resizebox{0.65\hsize}{!}{$L_{XTCL}=\mathbb{E}_{(\boldsymbol{x}^{l_i}, \boldsymbol{x}^{l_j})\sim D_{t}}\Tilde{L}_{xltcl}(\boldsymbol{x}^{l_i}).$}
\label{eq:objective_xltcl}
\end{equation}
In this way, the relevance of two arbitrary pieces of texts can be measured by the proximity of their TR representations in the UTRS, which will be used in the next pre-training objective.

\begin{figure*}[!htb]
\centering
\includegraphics[scale=0.60]{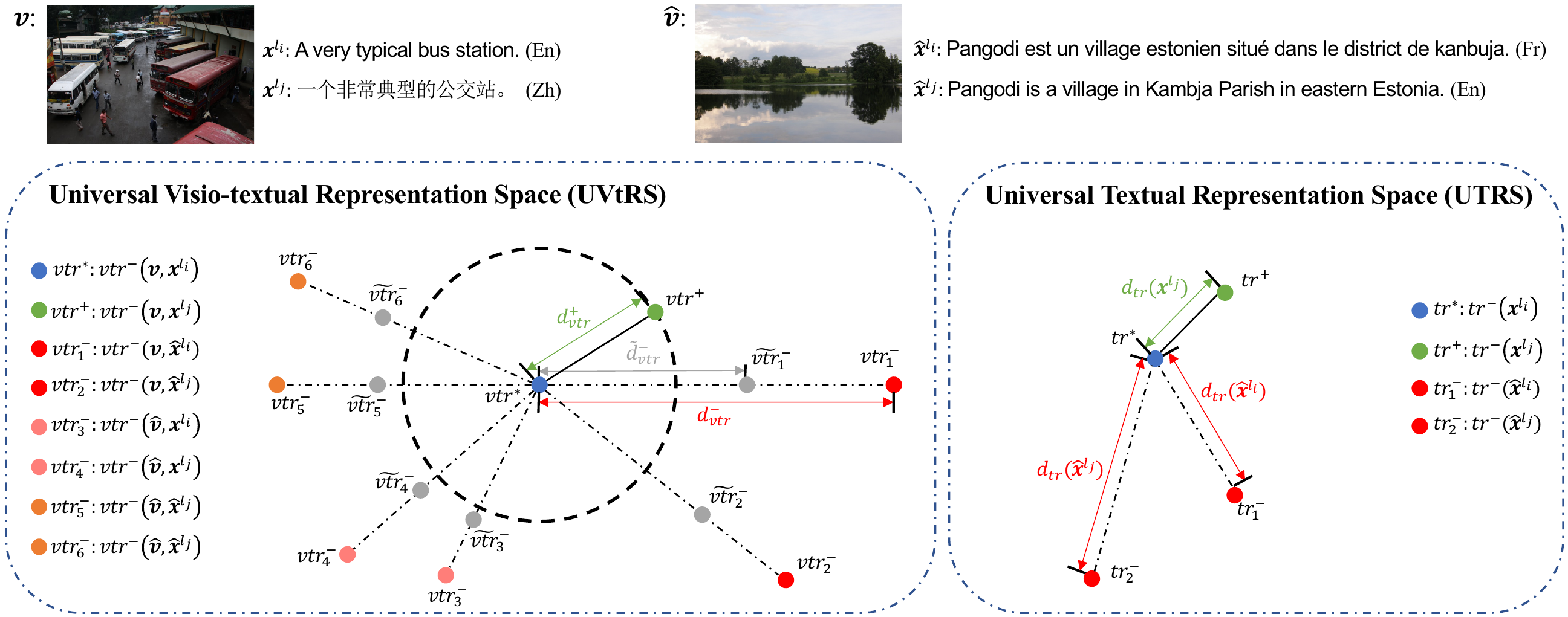}
\caption{
The illustration of Regularized Cross-lingual Visio-textual Contrastive Learning (R-XVtCL). In the UVtRS, the red, pink and brown points out of the circle correspond to VtR representations of the three types of negative samples. The grey ones are the interpolated VtR representation that are harder for the model to discriminate. In the UTRS, the distance between any two TR representations measures the textual relevance of their related texts.
}
\label{fig:objective_R-XVtCL}
\vspace{-10pt}
\end{figure*}
\subsubsection{Regularized Cross-lingual Visio-textual Contrastive Learning (R-XVtCL)}
\label{subsubsec:r-vtcl}
Similarly, the R-XVtCL objective is to learn semantically informative representations of visio-textual inputs in a universal visio-textual representation space (UVtRS), which involves both strictly-aligned and weakly-aligned image-caption pairs. We treat visio-textual inputs in another representation space because they differ from textual-only inputs in that their semantics depend on both images and texts. Analogously, we also expect the visio-textual representations (VtR) of semantically equivalent visio-textual inputs are near to each other. 

First, we introduce how to leverage the strictly-aligned multilingual image-caption pairs. Given a batch of image-caption triplets in two different languages $B_{vt}$=$\{(\boldsymbol{v}_b,\boldsymbol{x}^{l_{i}}_b,\boldsymbol{x}^{l_{j}}_b)\}_{b=1}^{|B_{vt}|}$ $(l_{i}$$\neq$$l_{j})$, for a triplet $(\boldsymbol{v},\boldsymbol{x}^{l_{i}},\boldsymbol{x}^{l_{j}})$$\in$$B_{vt}$, we use the pair $(\boldsymbol{v},\boldsymbol{x}^{l_{i}})$ as the anchor visio-textual instance, with its VtR representation $vtr^{*}$ serving as the anchor point in the UVtRS. Meanwhile, since $\boldsymbol{x}^{l_j}$ is parallel to $\boldsymbol{x}^{l_i}$, the pair $(\boldsymbol{v},\boldsymbol{x}^{l_{j}})$ is used as the positive sample, whose VtR representation $vtr^{+}$ should be close to $vtr^{*}$. Along with $(\boldsymbol{v},\boldsymbol{x}^{l_{i}},\boldsymbol{x}^{l_{j}})$, we construct three types of negative visio-textual samples using another triplet $(\hat{\boldsymbol{v}},\hat{\boldsymbol{x}}^{l_{i}},\hat{\boldsymbol{x}}^{l_{j}})$ within the same batch:
\begin{enumerate}
\setlength{\itemsep}{0pt}
\item[(1)] $(\boldsymbol{v},\hat{\boldsymbol{x}}^{l_i})$ and $(\boldsymbol{v},\hat{\boldsymbol{x}}^{l_j})$, containing the same image with the anchor instance but semantically non-equivalent captions;
\item[(2)] $(\hat{\boldsymbol{v}},\boldsymbol{x}^{l_i})$ and $(\hat{\boldsymbol{v}},\boldsymbol{x}^{l_j})$, containing semantically equivalent captions but different paired images;
\item[(3)] $(\hat{\boldsymbol{v}},\hat{\boldsymbol{x}}^{l_i})$ and $(\hat{\boldsymbol{v}},\hat{\boldsymbol{x}}^{l_j})$, containing different images and semantically non-equivalent captions.
\end{enumerate}
With these negative samples, we construct their harder representations in the UVtRS through the similar interpolation procedure described in Section~\ref{subsec:XTCL}, resulting in their interpolated VtR representations, as illustrated in Figure~\ref{fig:objective_R-XVtCL}.\footnote{We denote these VtR representations as $\widetilde{vtr}^{-}(\boldsymbol{v},\hat{\boldsymbol{x}}^{l_i})$, $\widetilde{vtr}^{-}(\boldsymbol{v},\hat{\boldsymbol{x}}^{l_j})$, $\widetilde{vtr}^{-}(\hat{\boldsymbol{v}},\boldsymbol{x}^{l_i})$, $\widetilde{vtr}^{-}(\hat{\boldsymbol{v}},\boldsymbol{x}^{l_j})$, $\widetilde{vtr}^{-}(\hat{\boldsymbol{v}},\hat{\boldsymbol{x}}^{l_i})$ and $\widetilde{vtr}^{-}(\hat{\boldsymbol{v}},\hat{\boldsymbol{x}}^{l_j})$.} Therefore, the contrastive loss using strictly-aligned multilingual image-caption pairs can be written as

\begin{small}
\begin{align}
&\Tilde{L}_{xlvtcl}(\boldsymbol{v},\boldsymbol{x}^{l_i}) = \\
&-\log{\frac{\exp{(-d^{+}_{vtr})}}{\exp{(-d^{+}_{vtr})}+\sum\limits_{(\boldsymbol{v}',\boldsymbol{x}') \in \mathcal{N}(\boldsymbol{v},\boldsymbol{x}^{l_i})}{\exp{(-\Tilde{d}^{-}_{vtr}(\boldsymbol{v}',\boldsymbol{x}'))}}}}, \notag
\end{align}
\end{small}where $\mathcal{N}(\boldsymbol{v},\boldsymbol{x}^{l_i})$ includes the above three types of negative samples, $d^{+}_{vtr}$ and $\widetilde{d}^{-}_{vtr}(\boldsymbol{v}',\boldsymbol{x}')$ are the euclidean distances from $vtr^{+}$ and each interpolated $\widetilde{vtr}^{-}(\boldsymbol{v}',\boldsymbol{x}')$ to the anchor in the UVtRS.

However, when using weakly-aligned multilingual image-caption pairs, it is not reasonable to simply encourage the VtR representation $vtr^{+}$ to be close to the anchor $vtr^{*}$ because $\boldsymbol{x}^{l_{i}}$ and $\boldsymbol{x}^{l_{j}}$ are not strictly parallel. Hence, we propose to constrain the proximity of $(\boldsymbol{v}, \boldsymbol{x}^{l_j})$ to the anchor instance $(\boldsymbol{v}, \boldsymbol{x}^{l_i})$ in the UVtRS through an additional regularization term, given that the proximity of two TR representations in the UTRS can be seen as textual relevance (See Section~\ref{subsec:XTCL}). 

Concretely, we first obtain the TR representations of all captions in the two weakly-aligned image-caption triplets $(\boldsymbol{v},\boldsymbol{x}^{l_{i}},\boldsymbol{x}^{l_{j}})$ and $(\hat{\boldsymbol{v}},\hat{\boldsymbol{x}}^{l_{i}},\hat{\boldsymbol{x}}^{l_{j}})$ from $D_w$. The textual relevances of $\boldsymbol{x}^{l_j}$, $\hat{\boldsymbol{x}}^{l_i}$ and $\hat{\boldsymbol{x}}^{l_j}$ with respect to $\boldsymbol{x}^{l_i}$ can be measured by the negative TR distance, \textit{i.e.} $-d_{tr}(\boldsymbol{x}^{l_j})$, $-d_{tr}(\hat{\boldsymbol{x}}^{l_i})$ and $-d_{tr}(\hat{\boldsymbol{x}}^{l_j})$, the closer to 0 the more relevant. Then, we transform these relevance scores into a normalized relevance distribution in the UTRS:
\begin{equation}
\resizebox{1.0\hsize}{!}{$P_{tr}={\rm softmax}([-d_{tr}(\boldsymbol{x}^{l_j}),-d_{tr}(\hat{\boldsymbol{x}}^{l_i}),-d_{tr}(\hat{\boldsymbol{x}}^{l_j})])$}.
\label{eq:tr_relevance}
\end{equation} 
Moreover, in the UVtRS, we can also obtain such a normalized relevance distribution $P_{vtr}$. Concretely, we select the image-text pairs that contain the same image as the anchor visio-textual instance $(\boldsymbol{v},\boldsymbol{x}^{l_i})$, including $(\boldsymbol{v},\boldsymbol{x}^{l_j})$, $(\boldsymbol{v},\hat{\boldsymbol{x}}^{l_i})$ and $(\boldsymbol{v},\hat{\boldsymbol{x}}^{l_j})$, because their VtR representation differences with the anchor only derive from semantically non-equivalent texts. Thereafter, $P_{vtr}$ can be computed as 
\begin{align}
P_{vtr}=&{\rm softmax}([-d_{vtr}(\boldsymbol{v},\boldsymbol{x}^{l_j}), \\
&-d_{vtr}(\boldsymbol{v},\hat{\boldsymbol{x}}^{l_i}),-d_{vtr}(\boldsymbol{v},\hat{\boldsymbol{x}}^{l_j})]). \notag
\end{align}
Hence, the regularized contrastive loss with weakly-aligned multilingual image-text pairs is:
\begin{equation}
\Tilde{L}^{reg}_{xlvtcl}(\boldsymbol{v},\boldsymbol{x}^{l_i}) = \Tilde{L}_{xlvtcl}(\boldsymbol{v},\boldsymbol{x}^{l_i})+{\rm KL}(P_{vtr}||P_{tr}).
\label{eq:xvtcl_reg}
\end{equation} 
Finally, with training instances from both $D_s$ and $D_w$, the R-XVtCL objective can be formulated as the following:
\begin{align}
L_{R-XVtCL}&=\mathbb{E}_{(\boldsymbol{v},\boldsymbol{x}^{l_i},\boldsymbol{x}^{l_j})\sim D_{s}}\Tilde{L}_{xlvtcl}(\boldsymbol{v},\boldsymbol{x}^{l_i}) + \notag \\
&\mathbb{E}_{(\boldsymbol{v},\boldsymbol{x}^{l_i},\boldsymbol{x}^{l_j})\sim D_{w}}\Tilde{L}^{reg}_{xlvtcl}(\boldsymbol{v},\boldsymbol{x}^{l_i}).
\label{eq:r-xvtcl}
\end{align}
Note that $D_s$ and $D_w$ are simultaneously used in this task. In particular, images from $D_s$ are processed into ROI-based visual features while those from $D_w$ are in the form of patch-based features. This is due to the fact that in general scenarios, the cost of obtaining ROI features of all images from much more abundant weakly-aligned image-text data is often unbearable.

\section{Experiments}
\subsection{Downstream Tasks}
We conduct experiments on five downstream multi-modal tasks across 6 languages (English, German, French, Indonesian, Japanese and Chinese), including Cross-lingual Visual Natural Language Inference (\textbf{XVNLI}), Cross-lingual Grounded Question Answering (\textbf{xGQA}), Multicultural Reasoning over Vision and Language (\textbf{MaRVL}), Image-Text Retrieval (\textbf{ITR}) and Multi-modal Machine Translation (\textbf{MMT}). The first four are discriminative tasks while the last one is a generative task. The details about these tasks and their datasets are given in Appendix~\ref{sec:downstream_tasks_datasets}.

\subsection{Implementation Details}
Following the setting of MBart-50~\cite{DBLP:journals/corr/abs-2008-00401}, our model consists of 12 encoder layers and 12 decoder layers with 16 attention heads and 1024 hidden dimensions, which is initialized by MBart-50 parameters. For visual inputs, the dimension of ROI-based features and patch embeddings are 2048 and 768, respectively. We use the ROI features provided in IGLUE~\cite{DBLP:conf/icml/Bugliarello0PRE22} generated from Faster-RCNN~\cite{NIPS2015_14bfa6bb}, which contain 36 regions for each image. Every original image is resized to 224$\times$224 pixels and then mapped to a flattened one-dimensional patch embedding sequence, where the patch size is set to 32$\times$32. For text inputs, we build a vocabulary out of the original one used in MBart-50, achieving a cover rate of over 99.99\% on the seven languages involved in our pre-training and downstream tasks. 

During pre-training, we use Adam optimizer~\cite{DBLP:journals/corr/KingmaB14} with a learning rate of $5$$\times$$10^{-5}$. We use DeepSpeed to support multi-node training. It takes about ten days to converge on 64 V100 GPUs, where the model is updated for 100,000 steps and the batch size is set to 1024. More details are given in Appendix~\ref{sec:downstream_tasks_datasets}.

\subsection{Contrast Models}
For the four discriminative tasks, we compare our model with recent competitive multilingual V\&L pre-trained models trained with strictly-aligned multilingual image-caption dataset: \textbf{M${^3}$P}~\cite{DBLP:conf/cvpr/NiHSCBW0D21}, \textbf{mUNITER}, \textbf{xUNITER}~\cite{DBLP:conf/emnlp/0001BPRCE21} and \textbf{UC$^2$}~\cite{DBLP:conf/cvpr/ZhouZW0LYL21}.
Meanwhile, we make comparison with several strong baselines, including \textbf{MeMAD}~\cite{DBLP:conf/wmt/GronroosHKLMPSS18}, \textbf{VL-T5} and \textbf{VL-BART}~\cite{DBLP:conf/icml/ChoLTB21}. All of these contrast models leverage ROI-based visual features during their pre-training and fine-tuning.

\begin{table}[]
\begin{center}
\small
\begin{tabular}{l|c|c|c|c} \hline
\textbf{Model}/\textbf{Task} & XVNLI & xGQA & MaRVL & ITR \\
\hline
\emph{M$^3$P} & 76.89 & 53.75 & 68.22 & 27.97 \\
\emph{mUNITER} & 76.38 & 54.68 & \textbf{71.91} & \textbf{42.70} \\
\emph{xUNITER} & 75.77 & 54.83 & 71.55 & 35.25 \\
\emph{UC$^2$} & 76.38 & 55.19 & 70.56 & 35.97 \\
\hdashline
\emph{RC$^3$-Patch} & 71.21 & 41.36 & - & - \\
\emph{RC$^3$-ROI} & 77.91 & 54.13 & 69.42 & 41.12 \\
\emph{RC$^3$-Combined} & \textbf{78.43} & \textbf{55.92} & 69.74 & 41.30 \\
\hline
\end{tabular}
\caption{Performances on English testsets of XVNLI, xGQA, MaRVL and ITR tasks. We report the average scores under three different random seeds.
}
\label{tab:results_english}
\end{center}
\vspace{-10pt}
\end{table}

\begin{table*}
\begin{center}
\begin{tabular}{l|c|ccc|cc|ccc}
\hline
\multirow{2}{*}{\textbf{Model}/\textbf{Task}} & \multicolumn{1}{c|}{XVNLI} & \multicolumn{3}{c|}{xGQA} & \multicolumn{2}{c|}{MaRVL} & \multicolumn{3}{c}{ITR} \\ \cline{2-10} 
 & \multicolumn{1}{c|}{Fr} & De & Id & \multicolumn{1}{c|}{Zh} & Id & \multicolumn{1}{c|}{Zh}   & De & Ja & \multicolumn{1}{c}{Zh} \\ \hline
\emph{M$^3$P} & 56.36 & 33.42 & 32.58 & 28.65 & 56.47 & 55.04 & 12.60 & 9.95 & 15.60 \\
\emph{mUNITER} & 59.36 &23.95 & 9.36 & 7.03 & 54.79 & 55.34 & 11.95 & 7.00 & 11.60 \\
\emph{xUNITER} & 63.32 & 34.83 & \textbf{33.73} & 19.55 & 55.14 & 53.06 & 13.95 & 10.50 & 15.87 \\
\emph{UC$^2$} & 69.67 & 42.85 & 28.67 & 31.16 & 56.74 & 59.88 & 26.25 & 23.32 & 28.95 \\
\hdashline
\emph{RC$^3$-Patch} & 64.43 & 24.44 & 22.53 & 25.97 & - & - & - & - & - \\
\emph{RC$^3$-ROI} & 71.65 & 40.39 & 29.24 & 36.06 & \textbf{57.80} & \textbf{62.55} & \textbf{35.20} & 30.82 & 35.52 \\
\emph{RC$^3$-Combined} & \textbf{72.43} & \textbf{43.69} & 31.94 & \textbf{39.49} & 57.26 &  60.77 & 34.35 & \textbf{30.90} & \textbf{37.20} \\
\hline
\end{tabular}
\caption {Zero-shot performances on non-English XVNLI, xGQA, MaRVL and ITR testsets. We also report the average scores under three different random seeds.
}
\label{tab:zeroshot_results}
\end{center}
\vspace{-10pt}
\end{table*}

\subsection{Evaluation on Discriminative Tasks}
In our experiments, we fine-tune the pre-trained model using only the English training data of each task and evaluate its performance on each target language, which means that the evaluations on non-English languages follow a zero-shot setting. The metrics of XVNLI, xGQA and MaRVL are accuracy and that of ITR is Recall@1. Note that there are two retrieval directions in ITR task: image-to-text and text-to-image, where the average Recall@1 on the two directions is reported in Table~\ref{tab:results_english} and Table~\ref{tab:zeroshot_results}. We denote our V\&L model trained using Patch-based, ROI-based and Combined visual features as \textbf{RC$^3$-Patch}, \textbf{RC$^3$-ROI} and \textbf{RC$^3$-Combined}, respectively. The reported results of other contrast models are provided in IGLUE benchmark~\cite{DBLP:conf/icml/Bugliarello0PRE22}.

\paragraph{Results on English Testsets.}
From Table~\ref{tab:results_english}, we can observe that \emph{RC$^3$-Combined} achieves better results on the English testsets of XVNLI and xGQA tasks over other contrast models, slightly underperforming \emph{mUNITER}, \emph{xUNITER} and \emph{UC$^2$} on MaRVL. Meanwhile, the ITR results of \emph{RC$^3$-ROI} and \emph{RC$^3$-Combined} surpass all other models except the best performing \emph{mUNITER}. Another phenomenon is that the inferiority of \emph{RC$^3$-Patch} over other variants indicates the importance of informativeness from visual features on these tasks, especially MaRVL and ITR where \emph{RC$^3$-Patch} is uncomparably worse. Whereas, \emph{RC$^3$-Combined} performs better than \emph{RC$^3$-ROI}, showing that additional patch embeddings still benefit the model to some extent. 

\paragraph{Zero-shot Results.}
Table~\ref{tab:zeroshot_results} gives the zero-shot performances on XVNLI, xGQA, MaRVL and ITR tasks across multiple non-English languages. Overall, we can see that our models, \emph{RC$^3$-ROI} and \emph{RC$^3$-Combined}, significantly outperform other contrast models. Particularly for ITR, the zero-shot results of our models exceed the strongest \emph{UC$^2$} model by considerable margins in all three languages. As for xGQA, though \emph{M$^3$P} and \emph{xUNITER} perform slightly better in Indonesian, our model \emph{RC$^3$-Combined} still achieves higher accuracy in German (43.69 \textit{v.s} 42.85) and especially Chinese (39.49 \textit{v.s} 31.16).

For MaRVL, it can be seen that although \emph{RC$^3$-Combined} surpasses other contrast models, it is inferior to \emph{RC$^3$-ROI}. We conjecture that this is due to the double-image nature of MaRVL task.\footnote{Please refer to Appendix~\ref{sec:downstream_tasks_datasets} for the details about MaRVL task and its specific visual input formats.} Concretely, when ``Combined'' visual features of the two involved images are fed together to the encoder, the excessive length of visual inputs might distract the model from adequately attending to the textual modality, which cannot offset the benefit gained from additional patch embeddings. Such effect particularly stands out in a zero-shot setting, where V\&L models more heavily rely on meaningful textual representations learned from pre-training and the English-only fine-tuning.
\begin{table}[]
\begin{center}
\small
\begin{tabular}{l|c|c|c|c} \hline
\multirow{2}{*}{\textbf{Model}/\textbf{Testset}} & \multicolumn{2}{c|}{En-De} & \multicolumn{2}{c}{En-Fr} \\ \cline{2-5}
 & 2016 & 2017 & 2016 & 2017\\ \hline
\emph{MeMAD} & 38.9 & 32.0 & 62.2 & 54.4 \\
\emph{VL-T5} & 45.5 & 40.9 & - & - \\
\emph{VL-BART} & 41.3 & 35.9 & - & - \\
\hdashline
\emph{RC$^3$-Patch} & 45.49 & \textbf{42.06} & 68.29 & 62.56 \\
\emph{RC$^3$-ROI} & 45.73 & 41.52 & 68.38 & \textbf{62.71} \\
\emph{RC$^3$-Combined} & \textbf{45.86} & 42.01 & \textbf{68.50} & 62.66 \\
\hline
\end{tabular}
\caption{Performances on Multi30k English-to-German (En-De) and English-to-French (En-Fr) testsets.
}
\label{tab:mmt_results}
\end{center}
\vspace{-10pt}
\end{table}

\begin{table*}
\begin{center}
\small

\scalebox{0.89}{\begin{tabular}{l|cc|cccc|ccc|cccc}
\hline
\multirow{2}{*}{\textbf{Model}/\textbf{Task}} & \multicolumn{2}{c|}{XVNLI} & \multicolumn{4}{c|}{xGQA} & \multicolumn{3}{c|}{MaRVL} & \multicolumn{4}{c}{ITR} \\ \cline{2-14} 
 & En & \multicolumn{1}{c|}{Fr} & En & De & Id & \multicolumn{1}{c|}{Zh} & En & Id & \multicolumn{1}{c|}{Zh} & En  & De & Ja & \multicolumn{1}{c}{Zh}\\ \hline
\emph{RC$^3$-Combined} & \textbf{78.43} & \textbf{72.43} & \textbf{55.92} & 43.69 & \textbf{31.94} & 39.49 & \textbf{69.74} & 57.26 & 60.77 & \textbf{41.30} & 34.35 & \textbf{30.90} & \textbf{37.25}\\
\emph{RC$^3$-ROI} & 77.91 & 71.65 & 54.13 & 40.39 & 29.24 & 36.06 & 69.42 & \textbf{57.80} & \textbf{62.55} & 41.12 & \textbf{35.22} & 30.82 & 35.55 \\
\hdashline
\emph{w/o.} ${\rm KL}(P_{vtr}||P_{tr})$ & 76.26 & 70.69 & 53.41 & 43.49 & 24.32 & 33.27 & 69.41 & 55.31 & 58.89 & 40.60 & 34.72 & 29.15 & 35.20 \\
\emph{w/o. R-XVtCL} & 74.34 & 70.26 & 52.63 & \textbf{44.87} & 20.06 & \textbf{41.93} & 69.28 & 50.62 & 57.41 & 40.82 & 34.02 & 30.72 & 35.02 \\
\emph{w/o. R-XVtCL\,\&\,XTCL} & 76.17 & 70.52 & 52.43 & 39.71 & 11.17 & 33.28 & 68.83 & 52.21 & 56.42 & 39.22 & 33.80 & 30.02 & 34.70 \\
\hline
\end{tabular}}

\caption {Ablation results. Note that all variants except \emph{RC$^3$-Combined} adopt ROI-based visual features for evaluation.}
\label{tab:ablation_study}
\end{center}
\vspace{-15pt}
\end{table*}

\subsection{Evaluation on MMT}
MMT is a generative task that involves both encoder and decoder to generate translations based on source sentences and their paired images. Table~\ref{tab:mmt_results} lists the performances on Mulit30K English-to-German (En-De) and English-to-French (En-Fr) datasets. We can see that our models outperform other contrast models. Nevertheless, according to previous research~\cite{DBLP:conf/naacl/CaglayanMSB19}, it shows that the source sentences in Multi30k dataset presumably take more effects than images for translations, which could explain the outcome that our three model variants exhibit no obvious differences.

\section{Ablation Study}
In this section, we conduct ablation studies to investigate the effect of our proposed training objectives in Section~\ref{subsec:pretrain_objectives}. Adopting ROI-based visual features, we investigate the following three model variants:
\vspace{-5pt}
\begin{itemize}
    \setlength{\itemsep}{-0pt}
    \item \emph{w/o.}~${\rm KL}(P_{vtr}||P_{tr})$: This variant removes the regularization term in Equation~\ref{eq:xvtcl_reg}, which means that the weakly-aligned multilingual image-caption pairs from $D_w$ are used in the same way as strictly-aligned ones.
    \item \emph{w/o.}~\emph{R-XVtCL}: In this variant, the R-XVtCL objective is totally removed during pre-training.
    \item \emph{w/o.}~\emph{R-XVtCL\,\&\,XTCL}: In this variant, we remove both XTCL and R-XVtCL objectives, only using MCLM and ITM for pre-training.
\end{itemize}
From Table~\ref{tab:ablation_study}, it is clear that the removal of ${\rm KL}(P_{vtr}||P_{tr})$ in Equation~\ref{eq:xvtcl_reg} gives rise to performance drops, which demonstrates the effectiveness of constraining the VtR representation proximity of multilingual weakly-aligned image-caption pairs. In Appendix~\ref{sec:case_study}, we give several illustrative cases that present how our proposed textual relevance-based regularization affects the VtR representation proximity in the UVtRS. Moreover, although \emph{w/o.}~\emph{R-XVtCL} achieves the highest accuracy on German and Chinese xGQA datasets, it still mostly underperforms compared to \emph{w/o.}~${\rm KL}(P_{vtr}||P_{tr})$, \emph{RC$^3$-ROI} and \emph{RC$^3$-Combined}. This shows that the R-XVtCL objective brings improvement to our model by enhancing the learned VtR representations. Besides, removing both R-XVtCL and XTCL results in worse performances compared to the other two ablation variants except on XVNLI.

\section{Related Work}
In recent years, there have been a series of V\&L pre-trained models achieving remarkable progress on many downstream multi-modal tasks. Overall, these studies adjust model architectures and design various pre-training objectives to learn alignment between visual and textual modalities. They can be mainly classified into single-stream~\cite{DBLP:conf/eccv/ChenLYK0G0020,DBLP:conf/icml/ChoLTB21,DBLP:conf/iclr/WangYYDT022} and two-stream V\&L architectures~\cite{DBLP:conf/nips/LuBPL19,DBLP:journals/corr/abs-2206-00621}.

Apart from the above models, some multilingual V\&L pre-trained models are proposed to learn universal representations across multiple languages and modalities. One of the major difficulties is the lack of high-quality multilingual multi-modal pre-training data. To address this issue, \citet{DBLP:conf/cvpr/NiHSCBW0D21} proposed to integrate multilingual pre-training and multi-modal pre-training. Concretely, batches of multilingual text corpora and monolingual multi-modal data are alternately used. Following a similar manner, \citet{DBLP:conf/emnlp/0001BPRCE21} build \emph{mUNITER} and \emph{xUNITER} by initializing model parameters with \emph{mBERT} and \emph{XLM-R}, respectively. Furthermore, \citet{DBLP:conf/cvpr/ZhouZW0LYL21} translate original English pre-training data into multiple languages and propose \emph{UC$^2$} to learn universal representations by introducing two cross-lingual/cross-modal pre-training tasks. These models leverage strictly-aligned multilingual and multi-modal datasets that are relatively difficult to collect. Therefore in this paper, we additionally make better use of more abundant weakly-aligned multilingual multi-modal data.

\section{Conclusion}
In this paper, we propose Regularized Contrastive Cross-lingual Cross-modal pre-training, which additionally exploits relatively more abundant weakly-aligned multilingual image-text pairs. During pre-training, we constrain the proximity of visio-textual representations of weakly-aligned image-text pairs according to their textual relevance. Besides, we further enhance our V\&L model by integrating ROI-based and patch-based visual features. Compared with recent competitive V\&L models, our model achieves higher or comparable results, especially demonstrating stronger zero-shot performance.

\section*{Limitations}
Currently, we build a vocabulary from the original one used in MBart-50, and only conduct downstream experiments across 6 languages (English, German, French, Indonesian, Japanese and Chinese). Although we could involve more languages, it would require a larger CUDA memory that might go beyond our device capacity. Hence, we merely select the above languages that have sufficient overlap with our pre-training datasets. In addition, for fair comparisons, we only use the strictly-aligned multilingual multi-modal dataset provided in~\cite{DBLP:conf/cvpr/ZhouZW0LYL21}, which is augmented through machine translation. It is unclear how the quality of strictly-aligned dataset would affect model performance. Meanwhile, the length of texts in our weakly-aligned multilingual multi-modal dataset is generally very long. As a result, we truncate textual inputs before feeding them into the encoder, possibly bringing information loss to some extent.


\bibliography{anthology,custom}
\bibliographystyle{acl_natbib}

\clearpage
\appendix

\begin{table*}
\begin{center}
\small
\begin{tabular}{l|c|c|c|c|c|c} \hline
 & En & De & Fr & Ja & Zh &Id \\
\hline
En & 5,157,134 & 739,697 & 814,485 & 376,759 & 357,677 & 163,442 \\\hline
De & 739,697  & 3,248,830 & 516,048 & 199,996 & 163,226 & 77,632 \\\hline
Fr & 814,485  & 516,048 & 2485,944 & 223,177 & 188,968 & 91,712 \\\hline
Ja & 376,759  & 199,996 & 223,177 & 1,032,183 & 174,226 & 67,030 \\\hline
Zh & 357,677  & 163,226 & 188,968 & 174,226 & 798,853 & 66,294 \\\hline
Id & 163,442  & 77,632 & 91,712 & 67,030 &66,294 & 266,144 \\
\hline
\end{tabular}
\caption{Detailed statistics of weakly-aligned multilingual image-text dataset $D_w$.}
\label{tab:wit-relation}
\end{center}
\end{table*}
\section{Pre-training Data}
\label{sec:pretrain_datasets}
As described in Section~\ref{subsection:pretrain_data}, our pre-training involves three types of data:

\paragraph{\textbf{Strictly-aligned Multilingual Image-caption Dataset} $D_{s}$.}
Following previous work~\cite{DBLP:conf/icml/Bugliarello0PRE22}, we use the ConceptCaption dataset as the strictly-aligned multilingual image-caption dataset $D_s$, which contains the original 2,777,649 image-caption pairs and machine-translated captions in five other languages (Czech, German, French, Japanese and Chinese). Besides, during pre-training, we use the pre-processed ROI features provided in IGLUE benchmark~\cite{DBLP:conf/icml/Bugliarello0PRE22}

\paragraph{\textbf{Weakly-aligned Multilingual Image-text Dataset $D_w$.}} This dataset is built from a fraction of the publicly-available WIT~\cite{wit-2022-deriving} dataset. In WIT, there are a large number of unique images that have multiple pieces of related texts in different languages. First, we index images through their unique urls. Then, each image is paired with multiple pieces of related texts of different languages, resulting in a multilingual image-text tuple $(\boldsymbol{v},\boldsymbol{x}^{l_i},\boldsymbol{x}^{l_j},...,\boldsymbol{x}^{l_k})$ that shares the same image. The statistics of the constructed weakly-aligned dataset is provided in Table~\ref{tab:wit-relation}, where each entry represents the number of multilingual image-text tuples in the corresponding language pair.

\paragraph{\textbf{Multilingual Parallel Text Dataset $D_t$.}}
For this dataset used in XTCL task, we combine the parallel texts from $D_s$ and a subset of WikiMatrix~\cite{schwenk-etal-2021-wikimatrix} used in~\cite{DBLP:journals/corr/abs-2206-00621}. As a result, $D_t$ contains multilingual parallel texts of 7 languages, covering all languages involved in the pre-training and all downstream tasks, \textit{i.e.} Czech, German, French, Indonesian, Japanese and Chinese. 

\begin{table}
\begin{center}
\small
\scalebox{0.90}{
\begin{tabular}{l|c|c|c} \hline
 Hyperparameters & XVNLI &xGQA  & MaRVL  \\\hline
Learning Rate & 4e-5 & 4e-5 & 4e-5  \\\hline
Batch size & 128  & 256 & 64  \\\hline
Epochs & 10  & 5 & 40 \\\hline
Input length & 80  & 40 & 80 \\\hline
Hyperparameters & ITR & MMT (En-De) & MMT (En-Fr)\\\hline
Learning Rate  & 1e-5 & 5e-6 & 5e-6 \\\hline
Batch size & 64 & 256 & 256 \\\hline
Epochs  & 10 & 5 & 5 \\\hline
Input length & 80 & 50 & 50 \\\hline
\end{tabular}}
\caption{Hyperparameters for downstream tasks.}
\label{tab:hyper}
\end{center}
\vspace{-5pt}
\end{table}
\section{Downstream Tasks and Datasets}
\label{sec:downstream_tasks_datasets}
We conduct experiments on five downstream multi-modal tasks: XVNLI, xGQA, MaRVL, ITR and MMT. For all downstream tasks, we fine-tune the model on English training sets, and then evaluate performances across all languages. The hyperparameters used in our experiments are listed in Table~\ref{tab:hyper}.

\paragraph{XVNLI.} Cross-lingual Visual Natural Language Inference task aims to discriminate whether a given textual hypothesis \textit{entails}, \textit{contradicts}, or is \textit{neutral} an image premise. Its dataset combines three existing text-only datasets SNLI~\cite{bowman-etal-2015-large}, with their cross-lingual~\cite{agic-schluter-2018-baselines} and multi-modal~\cite{DBLP:journals/corr/abs-1901-06706} counterparts.

\paragraph{xGQA.} The goal of Cross-lingual Grounded Question Answering task is to answer several types of structured questions about an image. The corresponding dataset is manually translated from the GQA~\cite{pfeiffer-etal-2022-xgqa} validation set into 7 languages.

\paragraph{MaRVL.} Multicultural Reasoning over Vision and Language task~\cite{liu-etal-2021-visually} requires the model to determine whether a textual description is true or false about a pair of images. Following~\cite{DBLP:conf/icml/Bugliarello0PRE22}, the NLVR2 dataset~\cite{suhr-etal-2019-corpus} is used for training while the MaRVL dataset is used for testing. Because the V\&L model needs to take in two images as inputs in this task, the input format of visual features is different from other tasks. Specifically, given a piece of text $\boldsymbol{x}$ and an image pair $(\boldsymbol{v}^1, \boldsymbol{v}^2)$, we concatenate visual and textual features as ${\rm [CLS]}$, $v^1_1$, $v^1_2$, ..., $v^1_k$, ${\rm [SEP']}$, $v^2_1$, $v^2_2$, ..., $v^2_k$, $\rm {[BOS]}$, $x_1$, $x_2$, $x_{|\boldsymbol{x}|}$, $[{\rm LAN}_{src}]$, where a special token ${\rm [SEP']}$ is inserted between two images. In the same way, the top-layer hidden state corresponding to ${\rm [CLS]}$ is used as the final visio-textual representation for fine-tuning and evaluation.

\paragraph{ITR.}
Image-Text Retrieval task is composed of image-to-text and text-to-image retrieval. Image-to-text retrieval is to select out the most relevant texts from a candidate set given an image. Inversely, text-to-image retrieval is to pick the most relevant image. We also use the ITR dataset provided in~\cite{DBLP:conf/icml/Bugliarello0PRE22}, which is collected by combining 1,000 images from Flickr30K~\cite{DBLP:journals/tacl/YoungLHH14} and 1,000 from MSCOCO~\cite{DBLP:conf/eccv/LinMBHPRDZ14}.

\paragraph{MMT.}
Multi-modal Machine Translation task is to translate a source sentence with the help of its paired image. We conduct experiments on the widely-used Multi30k dataset~\cite{elliott-etal-2016-multi30k}, where each image is paired with one English description and human translations into German\&French. The training and validation sets contain 29,000 and 1,014 instances, respectively. Besides, the test sets consist of \textit{test2016} and \textit{test2017}, each of which contains 1,000 instances for evaluation.



\begin{figure*}[!htb]
\centering
\includegraphics[scale=0.60]{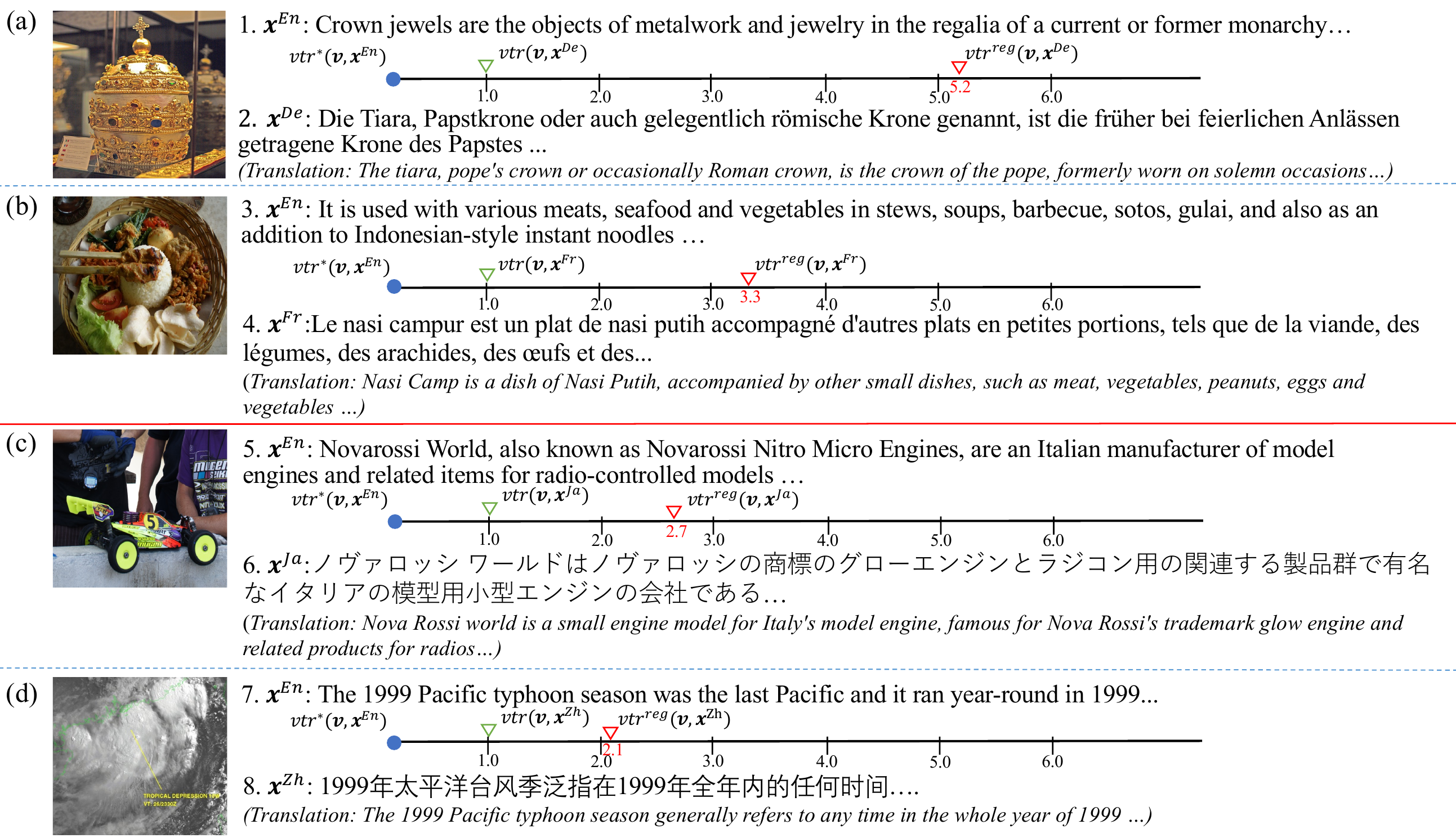}
\caption{Illustrative cases. For the axis of each case, the blue start point represents the anchor VtR representation. $vtr^{*}(\boldsymbol{v},\boldsymbol{x}^{En})$. The green positions on the axis represent the ratio unit 1.0, corresponding to the VtR representation without being regularized with ${\rm KL}(P_{vtr}||P_{tr})$. The red positions refer to the regularized VtR representation in terms of distance ratio in the UVtRS. 
}
\label{fig:case_study}
\vspace{-5pt}
\end{figure*}
\section{Case Study}
\label{sec:case_study}
In Figure~\ref{fig:case_study}, we exhibit several typical cases that can show the effect of our proposed regularization term ${\rm KL}(P_{vtr}||P_{tr})$ in Equation~\ref{eq:xvtcl_reg}, each of which contains an image and two pieces of texts. 
For each case, the image and its English texts are combined as the anchor visio-textual instance $vtr^{*}(\boldsymbol{v},\boldsymbol{x}^{En})$, corresponding to the blue start point in Figure~\ref{fig:case_study}. Similarly, the combination of the image and its non-English texts serves as the target visio-textual input whose euclidean VtR distance from $vtr^{*}(\boldsymbol{v},\boldsymbol{x}^{En})$ is worth probing. We introduce an axis to indicate the proximity of non-English visio-textual input to the anchor in the UVtRS with and without ${\rm KL}(P_{vtr}||P_{tr})$. 

Taking (a) for instance, let $vtr(\boldsymbol{v},\boldsymbol{x}^{De})$ and $vtr^{reg}(\boldsymbol{v},\boldsymbol{x}^{De})$ represent the VtR representations with and without regularization, respectively. We compute their euclidean distances to the anchor, denoted as $d_{vtr}$ and $d^{reg}_{vtr}$. Instead of marking the two absolute distances on the axis, we choose to record their ratio $d^{reg}_{vtr}/d_{vtr}$ that can reflect the proximity change after adding the regularization term ${\rm KL}(P_{vtr}||P_{tr})$. This is because the relative proximity is the what really matters for each case. Referring to the translations in italics, we can observe that the paired texts in cases (c) and (d) are more relevant to each other, \textit{i.e.} 1$\leftrightarrow$2 and 3$\leftrightarrow$4, than those in (a) and (b), \textit{i.e.} 5$\leftrightarrow$6 and 7$\leftrightarrow$8. Accordingly, it is clearly shown that the proximity changes of VtR representations are more significant in cases (a) and (b).

\end{document}